# Conditional Sparse Coding and Grouped Multivariate Regression

**Min Xu**                                                    MINX@CS.CMU.EDU

Machine Learning Department, Carnegie Mellon University, 5000 Forbes Avenue, Pittsburgh, PA 15213 USA

**John Lafferty**                                   LAFFERTY@GALTON.UCHICAGO.EDU

Department of Statistics and Department of Computer Science, University of Chicago, Chicago, IL 60637 USA

## Abstract

We study the problem of multivariate regression where the data are naturally grouped, and a regression matrix is to be estimated for each group. We propose an approach in which a dictionary of low rank parameter matrices is estimated across groups, and a sparse linear combination of the dictionary elements is estimated to form a model within each group. We refer to the method as *conditional sparse coding* since it is a coding procedure for the response vectors $Y$ conditioned on the covariate vectors $X$. This approach captures the shared information across the groups while adapting to the structure within each group. It exploits the same intuition behind sparse coding that has been successfully developed in computer vision and computational neuroscience. We propose an algorithm for conditional sparse coding, analyze its theoretical properties in terms of predictive accuracy, and present the results of simulation and brain imaging experiments that compare the new technique to reduced rank regression.

## 1. Introduction

Sparse coding, also called dictionary learning, is an approach to approximating a collection of signals by sparse linear combinations of a codewords chosen from a shared, learned dictionary. The method was proposed by Olshausen & Field (1996) for encoding natural images, with the motivation of developing a simple computational model of neural coding in the visual cortex. Through the use of sparsity and a large learned dictionary of codewords, sparse coding is able to efficiently capture a rich collection of features that are common to a population of signals. Variants of sparse coding have enjoyed considerable success in computer vision (Elad & Aharon, 2006; Lee et al., 2007; Mairal et al., 2009; Yang et al., 2009; Zhou et al., 2010; Bengio et al., 2009).

In this paper we apply the intuition behind sparse coding to design a new procedure for multivariate regression with data that fall into possibly overlapping groups or tasks. In traditional multivariate regression, the data consist of a set of response vectors $Y \in \mathbb{R}^q$, and for each $Y$, a corresponding covariate vector $X \in \mathbb{R}^p$. In a vector autoregressive time series model, for instance, $Y = Z_t$ is a vector at time $t$, and $X = Z_{t-1}$ is the vector at the previous step. In predicting brain activation patterns in neuroscience, $Y$ might be the neural activations in different regions of the brain with $X$ a vector of external stimuli. Under a linear model, $Y = BX + \epsilon$, where $B \in \mathbb{R}^{q \times p}$ is a matrix of parameters and $\epsilon \in \mathbb{R}^q$ is a random, mean zero error vector.

In many applications, the data naturally occur in groups or tasks, and assuming the same model $Y = BX + \epsilon$ for each group may be unjustified. For instance, in a non-stationary time series, the distribution of $Y = Z_t$ varies over time. In the neuroscience example, different people may have different neuronal activation patterns. In both cases it may be natural to place the data into possibly overlapping groups. More generally, the groups could be determined by any factor in the data or experimental design.

In settings where the input and output dimensions $p$ and $q$ are high, the number of parameters in $B$ may be be too large to estimate accurately from limited data. One approach to estimating reduced complexity models is to perform a least squares regression with a rank constraint on the coefficient matrix $B$. The nuclear norm serves as a convex surrogate for low rank constraints, and has be recently studied in the context of multivariate regression (Yuan et al., 2007; Negahban & Wainwright, 2011). For grouped data, a different model could be estimated for each group using this approach; however, carrying out separate regressions ignores commonality between the groups, and worsens the problem of limited data.

Our approach is to estimate the parameter matrices as

$$\widehat{B}^{(g)} = \sum_{k=1}^{K} \alpha_k^{(g)} D_k$$

where each dictionary entry $D_k$ is a low rank matrix, and $\alpha^{(g)} = (\alpha_1^{(g)}, \ldots, \alpha_K^{(g)})$ is a sparse vector; both $\{D_k\}$ and $\{\alpha^{(g)}\}$ are learned from data. The coefficients $\alpha_k^{(g)}$ are estimated for each group $g$, but the "codewords" or "dictionary elements" $D_k$ are shared across groups. This exploits the same intuition behind sparse coding for image analysis. Sparsity allows the dictionary entries $D_k$ to specialize and capture predictive aspects of the data shared by many groups, while the coefficients $\alpha^{(g)}$ tailor the model to the specific group $g$. Allowing the size $K$ of the dictionary to be large enables a rich class of parameter matrices to be modeled, while a low rank condition on the individual codeword matrices $D_k$ allows them to be estimated from limited data.

We perform both a "pessimistic" and "optimistic" analysis of our method. In the pessimistic analysis, the model may not be correct; that is, we do not assume any underlying common structure among the groups. In this case the model cannot achieve lower risk than the alternative of separate low rank regressions within each group. However, our analysis shows that the method suffers little excess risk relative to separate regressions. In the optimistic analysis, when the learned dictionary has captured common structure between the groups, the method produces an accurate estimator with much lower sample complexity than required by low rank regression. In both analyses, we measure statistical accuracy through non-asymptotic bounds on the excess risk $R(D, \alpha^{(g)}) - R(B^*)$. We show that the new procedure is effective and practical with experiments on simulated data and brain imaging data, reported in Section 6.

## 2. Related Work

Mairal et al. (2010) have studied a different way of using dictionary learning for supervised tasks; in this approach one first encodes data $X$ and then uses the encoding to perform classification or regression. Our work is more related to multi-task learning (Caruana, 1997; Evgeniou & Pontil, 2004) and is in particular a generalization of a model by Argyriou et al. (2006). They require that all $\alpha^{(g)}$ have the same sparsity pattern, so that all groups use the same small subset of dictionary elements. By allowing different groups to use different subsets of the dictionary, our model is much more flexible, though at the cost of requiring a non-convex optimization. Kang et al. (2011) used mixed integer programming to generalize the model of Argyriou et al. (2006) although our formulation is still more flexible and our optimization simpler. The approach of Liu et al. (2010) could be adapted to our setting, although their notion of task-relatedness is very different from ours.

Existing approaches to theoretical analysis of multitask learning differ significantly from our analysis by focusing on PAC-learnability with respect to a more abstract notion of task-relatedness (Maurer, 2006; Ben-David & Schuller, 2003). Theoretical analysis of sparse coding is rather limited. Some work studies the generalization error of dictionary learning (Vainsencher et al., 2010; Maurer & Pontil, 2010) and the local correctness of the non-convex objective for dictionary learning (Geng et al., 2011). Jeong & Kim (2009) consider sparse approximability and prove an information theoretic lower bound on sparse approximability of general $p$-dimensional vectors. They further show, non-constructively, that the lower bound can be achieved via an optimally constructed dictionary. We instead consider sparse approximability of a variety of structured spaces with respect to a dictionary that could plausibly be learned by a practical procedure.

## 3. Problem Formulation

In this work we focus on problems where the data are naturally grouped. Suppose we have $G$ groups, indexed by $g = 1, \ldots, G$. Let $X_i^{(g)} \in \mathbb{R}^p, Y_i^{(g)} \in \mathbb{R}^q$ denote the explanatory and response variables for the $i$th sample in group $g$. For each group, we let $B^{*(g)} = \arg\min_{B^{(g)}} R(B^{(g)})$ be the oracle regression matrix where we define

$$R(B^{(g)}) = \mathbb{E}_{X^{(g)}, Y^{(g)}} \|Y^{(g)} - B^{(g)} X^{(g)}\|_F^2.$$

For convenience, we will assume the sample size $n$ is the same for all groups, noting that more generally it will vary with $g$. Let $X^{(g)} = (X_1^{(g)}, \ldots, X_n^{(g)}) \in \mathbb{R}^{p \times n}$ and $Y^{(g)} = (Y_1^{(g)}, \ldots, Y_n^{(g)}) \in \mathbb{R}^{q \times n}$, with the $n$ samples of group $g$ arranged as matrix columns.

Our goal is to estimate $B^{*(g)}$. We consider estimates of the form $\widehat{B}^{(g)} = \sum_{k=1}^{K} \widehat{\alpha}_k^{(g)} D_k$ where each $D_k$ is a low rank matrix, and $\widehat{\alpha}^{(g)} = (\widehat{\alpha}_1^{(g)}, \ldots, \widehat{\alpha}_K^{(g)})$ is an estimated sparse vector. The codewords, or dictionary entries, $D_k$ are themselves estimated from data using nuclear norm regularization from data pooled across groups, as described in Section 4.

## 4. Conditional Sparse Coding

The basic idea underlying conditional sparse coding is to learn a collection of low rank matrices $\{D_1, ..., D_K\}$ (a dictionary) and estimate $\widehat{B}^{(g)}$ as a sparse linear combination of the dictionary entries. We optimize the overall objective function $f(\alpha, D)$ defined by

$$f(\alpha, D) =$$
$$\frac{1}{G} \sum_{g=1}^{G} \left\{ \frac{1}{n} \big\| Y^{(g)} - \Big( \sum_{k=1}^{K} \alpha_k^{(g)} D_k \Big) X^{(g)} \big\|_F^2 + \lambda \|\alpha^{(g)}\|_1 \right\}$$

where the optimization $\min_\alpha \min_{D \in \mathcal{C}_D(\tau)} f(\alpha, D)$ is carried out over the set

$$\mathcal{C}_D(\tau) = \left\{ D \in \mathbb{R}^{q \times p} : \|D\|_* \leq \tau \text{ and } \|D\|_2 \leq 1 \right\}.$$

The $\ell_1$ norm penalty induces sparsity on the $\alpha$ vectors and the nuclear-norm restriction forces the matrices $D_k$ to be low rank. The spectral norm constraint ensures no particular dictionary entry can be too large, and serves as an identifiability constraint; a similar constraint in sparse coding requires that all dictionary vectors must have norm no larger than one.

The objective function is biconvex but not jointly convex in $\alpha$ and $D$. Thus, we follow the standard sparse coding approach and alternately optimize over $\{\alpha^{(g)}\}$ with fixed $\{D_k\}$, and optimize over $\{D_k\}$ with fixed $\{\alpha^{(g)}\}$. We refer to the algorithm as *conditional sparse coding* (CSC) since it is a coding procedure for the response vectors $Y$ conditioned on the covariate vectors $X$.

---

**Algorithm 1** Conditional Sparse Coding (CSC)

Input: Data $\{(Y^{(g)}, X^{(g)})\}_{g=1,...,G}$, regularization parameters $\lambda$ and $\tau$.

1. Initialize dictionary $\{D_1, ..., D_K\}$ as random rank one matrices.

2. Alternate between the following steps until convergence of $f(\alpha, D)$:

   a. Encoding step: $\{\alpha^{(g)}\} \leftarrow \operatorname{argmin}_{\alpha^{(g)}} f(\alpha, D)$
   b. Learning step:
      $\{D_k\} \leftarrow \operatorname{argmin}_{D_k \in \mathcal{C}_D(\tau)} f(\alpha, D)$

---

The encoding step is equivalent to an independent $\ell_1$-constrained least squares fit, or lasso optimization, for each group $g$:

$$\min_{\alpha^{(g)} \in \mathbb{R}^K} \frac{1}{n} \sum_{i=1}^{n} \Big\| Y_i^{(g)} - \sum_{g=1}^{G} \alpha_k^{(g)} (D_k X_i^{(g)}) \Big\|_2^2 + \lambda \|\alpha^{(g)}\|_1. \tag{4.1}$$

A variety of algorithms are available to solve the lasso efficiently, notably iterative soft thresholding, a form of coordinate descent (Friedman et al., 2007).

For optimizing the dictionary entries, we designed both a projected gradient descent algorithm and a fast iterative shrinkage and thresholding algorithm (FISTA) following Beck & Teboulle (2009). A complication is that since the constraint set $\mathcal{C}_D(\lambda)$ is an intersection of nuclear norm and spectral norm balls, the projection needs to be done with care. We leave details of the optimization algorithms and the projection procedure to the appendix.

**Remarks on implementation details**
Although learning the dictionary is computationally intensive, fitting the coefficients to the dictionary is very fast due to efficient lasso optimization algorithms. Thus, an easy way to speed up CSC is to learn the dictionary with a smaller number of groups. The CSC optimization, being non-convex, is sensitive to initialization. We suggest random initialization both because our theoretical guarantees assume random initialization and because it works well in practice.

In sparse coding, one never picks a dictionary size $K$ equal to or greater than number of vectors to encode to avoid the trivial solution of letting each vector be a dictionary element itself. In CSC however, one can choose $K > G$ because of the nuclear-norm constraint on the dictionary entries. Based both on theory and experimental results, we recommend that $\tau$ is held to a constant between 1 and 0.5, and that $\lambda$ is then chosen with cross-validation.

## 5. Theoretical Analysis

To get a more complete understanding of CSC, we perform both a pessimistic analysis and an optimistic analysis. In the pessimistic analysis, we do not assume that our model is correct, and we do not assume any underlying common structure among the the groups. It is obvious that, under the general pessimistic setting, we cannot achieve higher statistical accuracy with CSC than with the alternative of estimating separate low-rank matrices for each group. Our pessimistic analysis provides a simple rule for determining, in the worst case, how much worse CSC is than the alternative.

In the optimistic analysis, we focus on a very specific setting where we only have to fit the coefficients to a pre-existing set of learned dictionary entries. We assume that the learned dictionary has thus captured common structure that exists among the groups. We show that in this setting CSC can produce an accurate estimator with fewer samples than the alternative of estimating separate matrices.

In all of our analyses, we measure statistical accuracy through non-asymptotic bounds on the excess risk $R(D, \alpha^{(g)}) - R(B^*)$. For clarity of presentation, we will use same symbols $c$ and $C$ to represent possibly different, generic constants in the theorem statements.

Before beginning the analysis, we enumerate and justify the underlying assumptions.

A1. For all groups $g$, $X^{(g)}$ and $Y^{(g)}$ are zero mean Gaussian random vectors. Let $\Sigma$ be the $(p+q) \times (p+q)$ covariance matrix $\Sigma = \mathbb{E}[(X^{(g)}, Y^{(g)})(X^{(g)}, Y^{(g)})^{\mathsf{T}}]$. Then the spectral norm $\|\Sigma\|_2$ is a constant independent of $n$.

A2. For all groups $g$, $\|B^{*(g)}\|_* \leq L$ and $B^{*(g)}$ is of rank at most $r$.

A3. The sample size satisfies $n \geq (p+q)$.

We make assumption A1 only to leverage results on concentration of measure; we do not use any other properties of the Gaussian distribution. Our analysis will thus easily extend to subgaussian random vectors. Assumption A2 is merely notation, allowing us to state our bounds in terms of $L$ and $r$. Assumption A3 is made so that many of the results in our pessimistic analysis can be stated more compactly; we do not make this assumption in our optimistic analysis.

It should be emphasized that since we are carrying out an excess risk analysis, we do not require incoherence conditions on our samples $X_1^{(g)}, \ldots, X_n^{(g)}$, as are often assumed in high-dimensional statistical analysis of sparsity.

Because we will repeatedly compare the excess risk rate of CSC against estimating separate matrices, we first prove an excess risk bound for using nuclear-norm regularization in each group.

**Theorem 5.1.** *Suppose that assumptions A1, A2, A3 hold. Let*

$$\widehat{B}^{(g)} = \underset{\{B : \|B\|_* \leq L\}}{\operatorname{argmin}} \frac{1}{n} \sum_{i=1}^{n} \|Y_i^{(g)} - BX_i^{(g)}\|_2^2.$$

*Then with probability at least $1 - \exp(-cp)$, we have that*

$$\max_{g=1,\ldots,G} R(\widehat{B}^{(g)}) - R(B^{*(g)}) \leq CL^2 \sqrt{\frac{(p+q)\log(nG)}{n}}$$

*where $c, C$ are constants depending only on $\|\Sigma\|_2$ as defined in A1.*

We provide proof sketches of all theorems in Section 5.3.

## 5.1. Pessimistic Analysis

Let $D^{\text{learn}}, \alpha_\lambda^{learn(g)}$ be the dictionary and coefficients output by Conditional Sparse Coding. The results of this section establish bounds on the excess risk $R(D^{\text{learn}}, \alpha_\lambda^{\text{learn}(g)}) - R(B^{*(g)})$. We stress that we do not assume $D^{\text{learn}}, \alpha_\lambda^{\text{learn}(g)}$ is the global minimizer of the non-convex CSC objective $f(\alpha, D)$. We use only the fact that the learned dictionary and coefficients achieve a lower objective than the random initial dictionary.

Before we state our main theorem, it is instructive to first consider the excess risk bound we would obtain if using only the random initial dictionary entries with oracle coefficients, with no additional dictionary learning.

**Proposition 5.1.** *Suppose that assumptions A1, A2, A3 hold. For a given sparsity level $s$, define*

$$\alpha_{oracle}^{init(g)} = \underset{\{\alpha^{(g)} : \|\alpha^{(g)}\|_0 \leq s, \|\alpha^{(g)}\|_1 \leq L\sqrt{s}\}}{\operatorname{argmin}} R(D^{init}, \alpha^{(g)}).$$

*Let $K \geq \max(n, r(p+q))$, and $\lambda \leq \sqrt{\frac{\log K}{n}}$. Suppose $s \leq r(p+q)$. Then with probability at least $1 - \frac{1}{K}$,*

$$\max_{g=1,\ldots,G} R(D^{init}, \alpha_{oracle}^{init(g)}) - R(B^{*(g)})$$

$$\leq CL^2 \left( \frac{(p+q)\log(GK)}{n} \right)^{s/r(p+q)}$$

*where $C$ is a constant depending only on $\|\Sigma\|_2$ as defined in A1.*

Setting $s = \frac{r(p+q)}{2}$, we observe that a large enough dictionary of random rank one matrices with the (non-sparse) oracle coefficients yields an excess risk bound that, up to multiplicative constants, matches the bound in Theorem 5.1—the best we can hope for. But because the oracle coefficients $\alpha_{\text{oracle}}^{\text{init}(g)}$ are not sparse, the learned coefficients $\alpha_\lambda^{\text{init}(g)}$ will be a poor estimate of the oracle coefficients, and the resulting excess risk may be significantly larger.

Proposition 5.1 and the preceding discussion motivate the need for learning the dictionary—we may improve statistical accuracy if we can customize the dictionary, allowing reconstruction of $B^{*(g)}$ from the dictionary using sparse coefficients. Our main theorem in this subsection formalizes this intuition.

**Theorem 5.2.** *Suppose assumptions A1, A2, A3 hold. Suppose $K \geq \max(n, r(p+q))$, $\lambda \leq \sqrt{\frac{\log K}{n}}$, and $\tau \leq 1$. Then with probability at least $1 - \frac{1}{K}$,*

$$\max_{g=1,\dots,G} R(D^{learn}, \alpha_\lambda^{learn}) - R(B^{*(g)})$$
$$\leq C \max(L^2, \|\alpha_\lambda^{learn}\|_1^2) \sqrt{\frac{(p+q)\log(GK)}{n}}.$$

This result implies that if the learned coefficients are sparse, that is, if $\|\alpha_\lambda^{\text{learn}(g)}\|_1 \leq L$, then the excess risk of conditional sparse coding is, up to a multiplicative constant factor, no greater than the excess risk for estimating separate low-rank matrices within each group. Of course, the excess risk can be worse if $\|\alpha_\lambda^{\text{learn}(g)}\|_1$ increases with $(p+q)$ or $n$; we cannot rule out this possibility because the dictionary learning optimization is nonconvex and does not admit a direct analysis. We note in our experimental section, however, that $\alpha_\lambda^{\text{learn}(g)}$ is very sparse in our simulations. We note also that our proof uses critically the fact that our algorithm places a nuclear-norm constraint on the dictionary entries, thus showing that the constraint is necessary to reduce overfitting when learning the dictionary.

Theorem 5.2 and Proposition 5.1 suggest a rule of thumb in applying conditional sparse coding. If the sparsity levels of the coefficients do not decrease with the iterations of dictionary learning, then the resulting statistical accuracy may be poor.

### 5.2. Optimistic Analysis

For our optimistic analysis, we consider the specific setting where the dictionary is already learned and we analyze the excess risk incurred when we fit the coefficients from data that were not used in the dictionary learning process.

A4. The learned dictionary $\{D^{\text{learn}}_1, \dots, D^{\text{learn}}_K\}$ is independent of the data $X_i^{(g)}$ for all groups $g$ and items $i = 1, \dots, n$.

With the dictionary fixed, we let

$$\alpha_{\text{oracle}}^{\text{learn}(g)} \equiv \underset{\{\alpha^{(g)} : \|\alpha^{(g)}\|_1 \leq L\}}{\arg\min} R(D^{\text{learn}}, \alpha^{(g)})$$

be the sparse coefficients that minimize the true risk. We can then interpret the oracle excess risk $R(D^{\text{learn}}, \alpha_{\text{oracle}}^{\text{learn}(g)}) - R(B^{*(g)})$ as a measure of the extent to which the oracle regression matrices $B^{*(g)}$ share structure, and the learned dictionary has captured this structure.

**Theorem 5.3.** *Suppose assumptions A1, A2, A4 hold. Suppose $\lambda \leq \sqrt{\frac{\log K}{n}}$. Then with probability at least $1 - \frac{1}{n}$,*

$$\max_{g=1,\dots,G} R(D^{learn}, \alpha^{learn(g)}) - R(B^{*(g)}) \leq$$
$$C \max(L^2, \|\alpha^{learn(g)}\|_1^2) \sqrt{\frac{\log(npKG)}{n}}$$
$$+ R(D^{learn}, \alpha_{oracle}^{learn(g)}) - R(B^{*(g)})$$

*where $C$ is some constant depending only on $\|\Sigma\|_2$ as defined in A1.*

Under the optimistic assumption that the excess risk $R(D^{\text{learn}}, \alpha_{\text{oracle}}^{\text{learn}(g)}) - R(B^{*(g)})$ is small, that is, that the dictionary has effectively learned the common information among the groups, then we require on the order of $\sqrt{p+q}$ times fewer samples here to achieve the same excess risk as in Theorem 5.2. If we further assume that $\|\alpha^{\text{learn}(g)}\|_1$ does not increase with $p$ and $q$, meaning that the oracle coefficients are sparse, then the excess risk in the optimistic setting is also lower than the bound in Theorem 5.1.

### 5.3. Proof Sketches

*Proof sketch of Theorem 5.1.* The crux of our argument is the following uniform generalization error bound.

**Lemma 5.1.** *With probability at least $1 - \exp(-cp)$, for all matrices $B^{(g)}$ such that $\|B^{(g)}\|_* \leq L$, $R(B^{(g)}) - \widehat{R}(B^{(g)}) \leq CL^2 \sqrt{\frac{(p+q)\log(Gn)}{n}} + R_u$, where $c$, $C$ are constants depending only on $\|\Sigma\|_2$ as defined in A1, and $R_u$ is a term that does not depend on $B^{(g)}$.*

We prove Lemma 5.1 by combining the technique of Greenshtein & Ritov (2004) with a concentration result from random matrix theory which states that for independent subgaussian random vectors $Z_1, \dots, Z_n$, $\|\frac{1}{n}\sum_{i=1}^n Z_i Z_i^\mathsf{T} - \Sigma_Z\|_2 \leq C\sqrt{\frac{p}{n}}$ with probability at least $1 - \exp(-cp)$ for some constants $c, C$. Theorem 5.1 then follows from a standard argument.

*Proof sketch of Proposition 5.1.* The proof is constructive. It uses a theoretical procedure, similar to orthogonal matching pursuit, but infeasible to implement, to produce a $\alpha^{(g)}$ with sparsity level $s$ for the random rank 1 dictionary entries so that the reconstruction error $\|B^{*(g)} - \sum_{k=1}^K D^{\text{init}} \alpha_k^{(g)}\|_F$ and the associated excess risk would be sufficiently low. Since $\alpha_{\text{oracle}}^{\text{init}}$ is the optimal set of $s$-sparse coefficients, we can upper bound its risk with the risk of our constructed coefficients. We do not prove that our bound is tight, but analysis by Jeong & Kim (2009) suggests that our bound cannot be significantly improved. We discuss this point further in the appendix.

*Proof sketch of Theorem 5.2.* We first rewrite the excess risk as

$$R(D^{\text{learn}}, \alpha_\lambda^{\text{learn}(g)}) - R(B^{*(g)})$$

$$= R(D^{\text{learn}}, \alpha_\lambda^{\text{learn}(g)}) - \widehat{R}(D^{\text{learn}}, \alpha_\lambda^{\text{learn}(g)}) \quad (5.1)$$

$$+ \widehat{R}(D^{\text{learn}}, \alpha_\lambda^{\text{learn}(g)}) - \widehat{R}(D^{\text{init}}, \alpha_{\text{oracle}}^{\text{init}(g)}) \quad (5.2)$$

$$+ \widehat{R}(D^{\text{init}}, \alpha_{\text{oracle}}^{\text{init}(g)}) - R(D^{\text{init}}, \alpha_{\text{oracle}}^{\text{init}(g)}) \quad (5.3)$$

$$+ R(D^{\text{init}}, \alpha_{\text{oracle}}^{\text{init}(g)}) - R(B^{*(g)}) \quad (5.4)$$

where $\alpha_{\text{oracle}}^{\text{init}(g)}$ is as defined in Proposition 5.1 with $s$ set to $\frac{r(p+q)}{2}$.

We then bound (5.1) using Lemma 5.1. To control (5.2), we observe that although the dictionary learning procedure is nonconvex, it is guaranteed to improve the objective. Thus, we have immediately that (5.2) is at most $\lambda \|\alpha_{\text{oracle}}^{\text{init}(g)}\|_1$. A bound on (5.4) follows from Proposition 5.1. Term (5.3) requires the following lemma concerning uniform generalization error of learning coefficients for a fixed dictionary.

**Lemma 5.2.** *Let $D_1, ..., D_K$ be a fixed set of dictionary entries with $\|D_k\|_* \leq 1$. We have that with probability at least $1 - \frac{1}{n}$, for all coefficients $\alpha^{(g)}$, $\max_g R(D, \alpha^{(g)}) - \widehat{R}(D, \alpha^{(g)}) \leq C \|\alpha^{(g)}\|_1^2 \sqrt{\frac{\log(GKpn)}{n}} + R_u$ where $C$ is a constant depending only on $\|\Sigma\|_2$ as defined in A1 and $R_u$ is a term that does not depend on $\alpha^{(g)}$*

*Proof sketch of Theorem 5.3.* The proof is straightforward by combining Assumption A4, Theorem 5.3, and Lemma 5.2.

# 6. Experiments

The main purpose of our experiments is to compare conditional sparse coding against reduced-rank regression. The experiments also illustrate that the coefficients estimated by CSC are indeed sparse and that the dictionary entries are low rank.

## 6.1. Simulation Data

We generate data using a linear model $Y^{(g)} = B^{*(g)}X^{(g)} + \epsilon^{(g)}$ where $\epsilon^{(g)} \sim N(0, \sigma^2 I_q)$ and each $B^{*(g)}$ is a $p \times p$ square matrix. We build a random design matrix $X^{(g)}$ by drawing each sample $X_i^{(g)} \sim N(0, I_p)$. We consider three different settings:

1. In the **structured** case, we construct each $B^{*(g)}$ as a random 3-sparse linear combination of a set

of 30 rank one dictionary matrices. Groups constructed by this method will share considerable common information; but, of course, the estimator has no knowledge of the true dictionary.

2. In the **unstructured** case, we construct each $B^{*(g)}$ as simply a random rank 3 matrix.

3. The **structured same design** case is the same as the structured case except that every group shares the same design $X^{(g)}$. We study this case because real-world data can have overlapping groups.

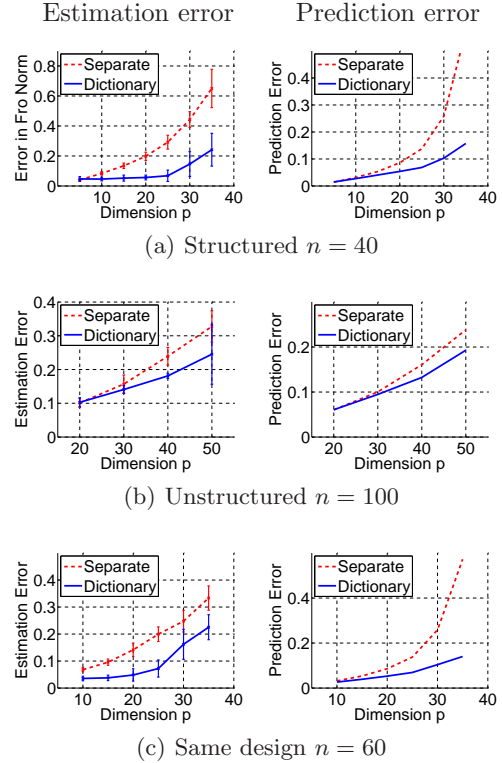

Estimation error    Prediction error

(a) Structured $n = 40$

(b) Unstructured $n = 100$

(c) Same design $n = 60$

*Figure 1.* Comparison of CSC to reduced rank regression.

We measure performance of the algorithms in terms of both *estimation error* $\frac{1}{G}\sum_{g=1}^{G} \|B^{*(g)} - \widehat{B}^{(g)}\|_F$ and *prediction error* $\widehat{R}_{test}(\widehat{B}^{(g)})$, which is computed from a large test set of $(X^{(g)}, Y^{(g)})$ pairs. We compare CSC against performing separate reduced rank regressions for each group using nuclear norm-regularization.

It can be seen from Figure 6.1 that when the parameter matrices $\{B^{*(g)}\}$ have significant common structure, CSC easily outperforms separate regressions with either different or the same design for each group. CSC performs worse in the unstructured case, as expected, but is still competitive with separate regressions.

In Figure 6.1, we show the sparsity of the coefficients together with the ranks of the learned dictionary en-tries, as a function of iterations of alternation in the algorithm. It is seen that (1) CSC does not require many iterations to converge, (2) the coefficients become increasingly sparse, and (3) although the ranks of the dictionary entries may increase, the learned dictionary entries are still relatively low rank.

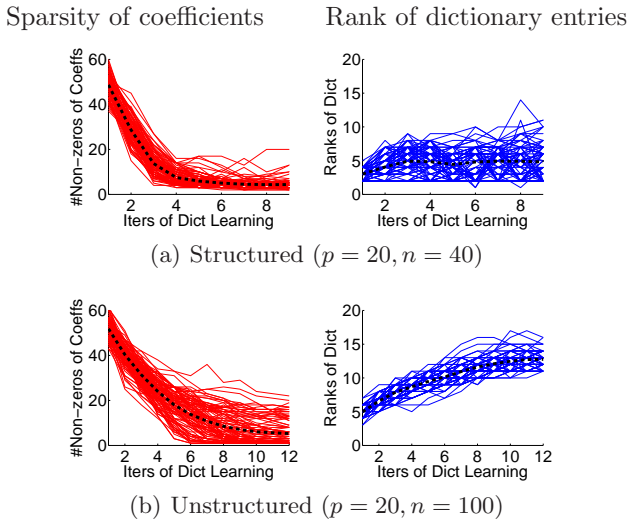

Sparsity of coefficients      Rank of dictionary entries

(a) Structured ($p = 20$, $n = 40$)

(b) Unstructured ($p = 20$, $n = 100$)

*Figure 2.* Variation in sparsity and dictionary rank with iterations of alternation in CSC dictionary learning, on simulated data. Each line represents one group or one dictionary entry; the dashed black line is the average.

We note that in Section 8 of the Appendix, we also perform simulations with overlapping groups.

### 6.2. fMRI Data

The dataset, gathered by Mitchell et al. (2008), comprises the brain activity patterns of 9 human subjects when presented with a single concrete English noun. We down-sample the original neural signal by retaining only one measurement in every $4 \times 3 \times 4$ voxel region of the brain. More precisely, we have $X$ as a design matrix of neural signals with dimension $(p = 434) \times (n = 60)$ and $Y$ as the response matrix with dimension $(q = 192) \times (n = 60)$ of semantic features of the 60 nouns being shown to the subjects. We let each subject be a group and hence we have that $G = 9$.

The goal is to predict the semantic features of the noun being shown to the subject, based only on the neural signal of the subject's brain. The predicted semantic features can then be used to guess which word the subject was viewing and thus "read the subject's mind."

Following Mitchell et al. (2008), we use hold-two-out cross-validation for evaluation. In each run of the experiment, we hold out two words, using the remaining

58 words for training, and then compute three evaluation metrics: 2 vs. 2 classification, 1 vs. 2 classification, and squared error. Let $y_1, y_2$ be the semantic feature vectors of the heldout words. Let $\widehat{y}_1, \widehat{y}_2$ be the predicted semantic feature vectors. We say that we correctly made a 2 vs. 2 classification if $d(y_1, \widehat{y}_1) + d(y_2, \widehat{y}_2) < d(y_1, \widehat{y}_2) + d(y_2, \widehat{y}_1)$ and we say that we correctly made a 1 vs. 2 classification if both $d(y_1, \widehat{y}_1) < d(y_1, \widehat{y}_2)$ and $d(y_2, \widehat{y}_2) < d(y_2, \widehat{y}_1)$. If we make random predictions, then the expected 1 vs. 2 classification accuracy is 0.25 and the expected 2 vs. 2 classification accuracy is 0.5. Our parameters are tuned by separate cross-validation trials. We used $K = 20$ dictionary entries.

In Figure 3, we compare the performance of CSC to separate trace-norm-regularized regressions for each subject. CSC often shows significant improvement in both 2 vs. 2 and 1 vs. 2 classification tasks, with very few cases of significant degradation. In terms of squared error, CSC shows improvement for most subjects, although on average, the improvement is statistically insignificant.

Although there is indeed sharing of dictionary entries across the various groups (subjects), it is important to mention that the pattern of sharing is unstable from trial to trial. Figure 6.2 shows two patterns of group-dictionary utilization derived from the $\alpha^{(g)}$ coefficients. We see that in the first trial, subject 3 shares significantly with subject 7, while subject 1 shares with no other subjects; in the second trial, subject 3 shares with subject 5 and subject 1 shares with subjects 6 and 9. The instability is possibly due to the low sample size. As a result of this instability, we cannot deduce subject-subject similarities from the dictionary utilization patterns.

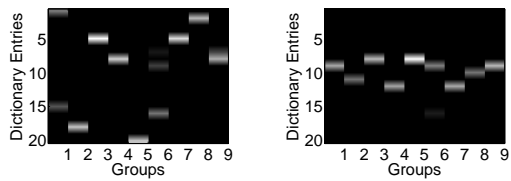

*Figure 4.* Coefficients $\alpha^{(g)}$ interpreted as dictionary utilization per group for two runs of the experiment. Lighter color indicates greater utilization.

### Acknowledgements

Research supported in part by NSF grant IIS-1116730 and AFOSR contract FA9550-09-1-0373.

|            | Subj A | B        | C        | D        | E        | F        | G        | H        | I        |
|------------|--------|----------|----------|----------|----------|----------|----------|----------|----------|
| Dictionary | 0.8833 | **0.8667** | 0.9000 | **0.9333** | **0.8333** | 0.7500 | **0.9000** | **0.7833** | **0.6667** |
| Separate   | **0.9500** | 0.7000 | **0.9167** | 0.8167 | 0.8167 | **0.7667** | 0.8000 | 0.6667 | 0.6333 |
| *Confidence* | 0.6- | 0.92+ | 0.05- | 0.86+ | 0.03+ | 0.02- | 0.70+ | 0.65+ | 0.07- |

|            | Subj A | B        | C        | D        | E        | F        | G        | H        | I        |
|------------|--------|----------|----------|----------|----------|----------|----------|----------|----------|
| Dictionary | 0.7000 | **0.5333** | 0.6667 | **0.7667** | 0.4833 | 0.3667 | **0.6000** | 0.4333 | **0.3000** |
| Separate   | 0.7000 | 0.3833 | 0.6667 | 0.6333 | **0.5000** | **0.5000** | 0.5333 | 0.3167 | 0.2333 |
| *Confidence* | 0 | 0.75+ | 0 | 0.72+ | 0.01- | 0.67- | 0.24+ | 0.58+ | 0.29+ |

|            | Subj A | B        | C        | D        | E        | F        | G        | H        | I        |
|------------|--------|----------|----------|----------|----------|----------|----------|----------|----------|
| Dictionary | **243.25** | **276.95** | **256.10** | **247.48** | **291.48** | **310.12** | 282.77 | 329.08 | 327.27 |
| Separate   | 255.00 | 290.94 | 270.30 | 253.73 | 299.84 | 322.59 | **272.45** | **314.69** | **303.41** |

*Figure 3.* fMRI data analysis. From top to bottom: 2v2 classification accuracy, 1v2 classification accuracy, and squared error from 60 cross-validation trials. Last row shows confidence in either improvement (+) or degradation (-).